\documentclass[10pt,twocolumn,letterpaper]{article}

\usepackage[pagenumbers]{cvpr}      
\definecolor{cvprblue}{rgb}{0.21,0.49,0.74}
\usepackage[pagebackref,breaklinks,colorlinks,allcolors=cvprblue]{hyperref}
\usepackage{graphicx}
\usepackage{cuted}
\usepackage{capt-of}
\usepackage{makecell}
\usepackage{comment}
\usepackage[accsupp]{axessibility}

\def\paperID{*****} 
\def\confName{CVPR}
\def\confYear{2026}

\title{An Approach to Enriching Surgical Video Datasets for Fine-Grained Spatial-Temporal Understanding of Vision-Language Models}

\author{Lennart Maack, Alexander Schlaefer  \\
Hamburg University of Technology, Germany \\
{\tt\small lennart.maack@tuhh.de}
}

\begin{document}
\maketitle
\begin{strip}
    \centering
    \includegraphics[width=\textwidth]{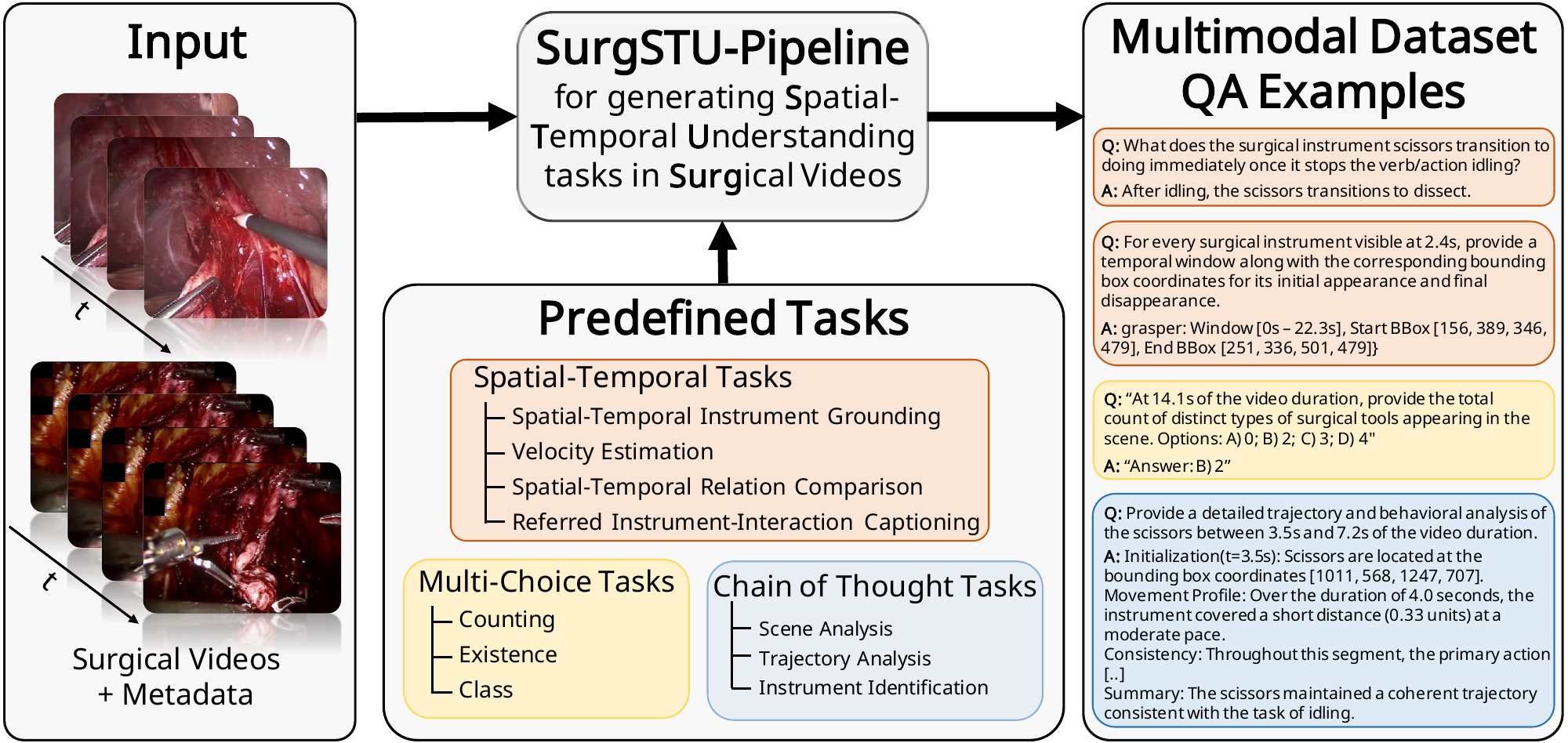}
    \captionof{figure}{The proposed SurgSTU-Pipeline for generating spatial-temporal understanding tasks leverages surgical videos with optionally available metadata, e.g. instrument-tissue interaction- or instrument localization information. By using predefined spatial-temporal task templates, the SurgSTU pipeline creates diverse Question-Answer pairs for training and evaluating Vision-Language Models.}
    \label{fig:dataset_curation_and_qa_generation}
\end{strip}

\begin{abstract}
Surgical video understanding is a crucial prerequisite for advancing Computer-Assisted Surgery.
While vision-language models (VLMs) have recently been applied to the surgical domain, existing surgical vision-language datasets lack in capturing and evaluating complex, interleaved spatial-temporal dynamics. 
Creating large scale datasets that accurately represent fine-grained spatial-temporal relationships in surgical videos is challenging due to costly manual annotations or error-prone generation using large language models. \\
To address this gap, we introduce the SurgSTU-Pipeline, a deterministic generation pipeline featuring temporal and spatial continuity filtering to reliably create surgical datasets for fine-grained spatial-temporal multimodal understanding.
Applying this pipeline to publicly available surgical datasets, we create the SurgSTU dataset, comprising 6711 video clips densely extended with 150k fine-grained spatial-temporal question-answer samples. \\
Our comprehensive evaluation shows that while state-of-the-art generalist VLMs struggle in zero-shot settings, their spatial-temporal capabilities can be improved through in-context learning.
A fine-tuned VLM on the SurgSTU training dataset achieves highest performance among all spatial-temporal tasks, validating the dataset's efficacy to improve spatial-temporal understanding of VLMs in surgical videos. The project is available here:
\mbox{\footnotesize \url{https://lennart-maack.github.io/SurgSTU-project/}}

\end{abstract}    
\section{Introduction}
\label{sec:intro}

Surgical scene understanding is crucial for enhancing Computer-Assisted Surgery (CAS) systems. These systems can support surgical training or improve robotic decision-making \cite{maier2017surgical, maier2022surgical, mascagni2022computer, ward2021computer}.
Due to their capabilities to generalize across multiple surgical scene understanding tasks, such as anatomy recognition, instrument detection, workflow recognition or surgical visual question answering, surgical vision-language models have demonstrated great potential \cite{zhou2023text, wang2024video, bai2023surgical, bai2025surgical, yuan2024advancing, seenivasan2023surgicalgpt}. \\
Training surgical vision-language models requires multimodal data, for example, image-text pairs. Early datasets enabled spatial grounding within question-answer frameworks by integrating bounding box annotations \cite{seenivasan2022surgical}. 
Other work utilized GPT-4V to address limitations in scale and language diversity, and expanded multimodal data for instruction-following from publicly available surgical datasets \cite{wang2025endochat}.
However, these datasets are limited to static frames and restricted in capturing the complex temporal dependencies and procedural context inherent in surgical videos. \\
To address this limitation, further work on vision-language models developed methods and datasets from videos to understand surgical procedures and their complex temporal dependencies beyond image-based static visual cues \cite{PEREZ2026103982, yuan2025learning, wang2025surgvidlm}. 
These datasets are often based on automatically transcribed, general narrations from surgical videos, which have the following limitations.
First, the transcribed text fails to accurately reflect the corresponding visuals, or it contains information irrelevant to surgical workflows.
Second, automatically transcribed labels from general narrations are still too coarse to capture the fine-grained structure of surgical activities and instrument handling.
However, accurately capturing the spatial-temporal dynamics of surgical instruments and interactions via vision-language models is essential for developing surgical training systems or enabling partial automation, e.g., through vision language action models. \\
Creating large scale datasets that accurately represent such fine-grained spatial-temporal relationships in surgical videos is challenging.
While manual annotation for fine-grained spatial-temporal relationships is significantly more costly than simple surgical video narration, automating large-scale question-answer (QA) generation via large language models introduces the risk of hallucinations and flawed ground truth. \\
Therefore, we propose SurgSTU-Pipeline, a deterministic pipeline with temporal and spatial continuity filtering to reliably create surgical datasets for fine-grained spatial-temporal multimodal understanding.
The SurgSTU pipeline first generates a unified fine-grained context tuple format by utilizing optionally available metadata from surgical datasets, such as information on frame-level instrument-tissue interactions and the localization of instruments.
Next, the pipeline uses the fine-grained context to create spatial-temporal instrument motion tracks using a continuity filter to reduce the propagation of annotation noise.
By using predefined task templates, the pipeline can generate fine-grained spatial-temporal question-answer samples. \\
To demonstrate the applicability of the proposed pipeline, we create the SurgSTU dataset and benchmark. 
The SurgSTU dataset and benchmark is created using publicly available surgical datasets of cholecystectomy and radical prostatectomy procedures.
To deterministically evaluate multimodal large language models for fine-grained spatio-temporal tasks in the SurgSTU dataset, we use a rule-based evaluation scheme.
We conduct a comprehensive baseline evaluation using state-of-the-art proprietary and open-source multimodal models, i.e., Gemini3.1 Flash-Lite and Qwen3-VL 2B, as well as a finetuned Qwen3-VL 2B model.
For the generalist models, we investigate the influence of in-context learning, i.e., providing domain-specific examples in the prompt.
Our results demonstrate that while generalist vision-language models initially struggle with dynamic spatial-temporal tasks, both in-context learning and domain-specific fine-tuning on the SurgSTU dataset enhance their fine-grained spatial-temporal reasoning capabilities. 
With our work, we establish the SurgSTU-Pipeline as a robust framework for generating fine-grained spatial-temporal datasets and benchmarks in surgery, addressing a critical gap in existing surgical multimodal datasets.

\section{Related Work}

\begin{table*}[t]
\centering
\caption{Comparison of recent multimodal surgical datasets. The SurgSTU-pipeline enables the generation of interleaved spatial-temporal tasks in surgical videos. Annotation Scope defines the dataset's complexity: ranging from basic visual summaries (Frame/Video-Level Description) to precise spatial localization (Grounding) and complex action tracking over time (Temporal Reasoning).}
\label{tab:dataset_comparison}
\resizebox{\textwidth}{!}{%
\begin{tabular}{llcccccc}
\toprule
\textbf{Dataset} & 
\textbf{\begin{tabular}[c]{@{}l@{}}Visual \\ Modality\end{tabular}} & 
\textbf{\begin{tabular}[c]{@{}c@{}}Number of \\ Images/ \\ Videos\end{tabular}} & 
\textbf{\begin{tabular}[c]{@{}c@{}}Number of \\ Visual-Text/\\ QA pairs\end{tabular}} & 
\textbf{\begin{tabular}[c]{@{}c@{}}Spatial \\ Grounding\end{tabular}} & 
\textbf{\begin{tabular}[c]{@{}c@{}}Temporal \\ Reasoning\end{tabular}} & 
\textbf{\begin{tabular}[c]{@{}c@{}}Spatial- \\ Temporal \\ Tasks\end{tabular}} & 
\textbf{\begin{tabular}[c]{@{}c@{}}Annotation Scope\end{tabular} } \\ \midrule

EndoVis-18-VQLA \cite{bai2023surgical} & Image & 2k     & 11.8k  & \checkmark & $\times$    & $\times$   & \begin{tabular}[c]{@{}c@{}}Frame-Level \\ Description $\&$ Grounding\end{tabular} \\ \cmidrule[0.2pt](lr){1-8}
SVL \cite{yuan2025learning}             & Video & 1.4k   & 25k & $\times$          & \checkmark & $\times$          & \begin{tabular}[c]{@{}c@{}}Video-Level \\ Description\end{tabular} \\ \cmidrule[0.2pt](lr){1-8}
Surg-396K \cite{wang2025endochat}       & Image & 41k & 396k & \checkmark & $\times$          & $\times$           & \begin{tabular}[c]{@{}c@{}}Multi-Level \\ Description $\&$ Grounding\end{tabular}  \\ \cmidrule[0.2pt](lr){1-8}
SVU-31K \cite{wang2025surgvidlm}         & Video & 1.9k & 31k & $\times$          & \checkmark& $\times$           & \begin{tabular}[c]{@{}c@{}}Multi-Level+Temporal \\ Description $\&$ Reasoning\end{tabular} \\ \cmidrule[0.2pt](lr){1-8}
SurgLaVi \cite{PEREZ2026103982}        & Video & 5.3k & 240k & $\times$          & \checkmark & $\times$           & \begin{tabular}[c]{@{}c@{}}Multi-Level+Temporal \\ Description $\&$ Reasoning\end{tabular} \\ \cmidrule[0.2pt](lr){1-8}
SurgSTU (Ours) & Video & 6.7k & 150k & \checkmark & \checkmark & \checkmark & \begin{tabular}[c]{@{}c@{}}Fine-Grained Spatial-Temporal \\ Description $\&$ Grounding\end{tabular} \\ \bottomrule
\end{tabular}%
}
\end{table*}

\label{sec:formatting}

\subsection{Surgical Vision-Language Models}

Recent advancements in surgical computer vision have increasingly adopted vision-language integration to overcome the limitations of strictly supervised, vision-only methods.
Early efforts focused on structured Visual Question Answering (VQA).
For example, Surgical-VQA and Surgical-VQLA facilitated scene understanding and spatial localization by predicting answers and bounding boxes simultaneously\cite{seenivasan2022surgical, bai2023surgical}.
However, these methods relied on rigid encoder-decoder architectures and predefined query formats, fundamentally limiting their ability to handle open-ended interactions and complex conversational flows. \\
In order to achieve broader generalization, current research has focused on two different paradigms: learning vision-language representations (pre-training) and generative multimodal large language models (MLLMs).
Representation learning methods aim to align visual and textual modalities into a shared, coordinated latent space, typically utilizing contrastive learning objectives.
This leads to features that excel at zero-shot discriminative tasks (e.g., surgical phase, tool, or action triplet recognition).
For instance, SurgVLP introduced a dual-branch contrastive framework that aligns video clips with automated transcriptions within a joint embedding space \cite{yuan2025learning}.
Building on this, HecVL proposed a hierarchical fine-to-coarse contrastive learning strategy that explicitly separates embedding spaces for atomic actions, phase-level summaries, and video-level abstracts \cite{yuan2024hecvl}.
Similarly, SurgCLIP utilizes a CLIP-style dual-encoder architecture coupled with dynamic temporal sampling to align multi-scale surgical clips with semantically rich captions\cite{PEREZ2026103982}.
These approaches maximize representational value from limited supervision but are inherently constrained by their dual-branch nature, restricting their application to generative tasks. \\
In contrast, generative methods leverage the profound reasoning and auto-regressive text generation capabilities of MLLMs to process unstructured visual cues and handle open-ended, multi-turn dialogues.
Models such as Surgical-LLaVA adapted general-purpose MLLMs to the surgical domain through visual instruction tuning \cite{jin2024surgical}.
EndoChat advanced this by introducing a grounded MLLM tailored for endoscopic surgery, employing a Mixed Visual Token Engine (MVTE) to extract multi-scale visual features and support diverse conversational types \cite{wang2025endochat}.
For more advanced temporal reasoning, SurgVidLM extends LLM capabilities to surgical video understanding via a two-stage StageFocus mechanism.
Multi-frequency visual tokens are integrated into the language model, such that global video context can be refined into fine-grained local analysis \cite{wang2025surgvidlm}.

\begin{figure*}[t]
    \centering
    \includegraphics[width=0.9\textwidth]{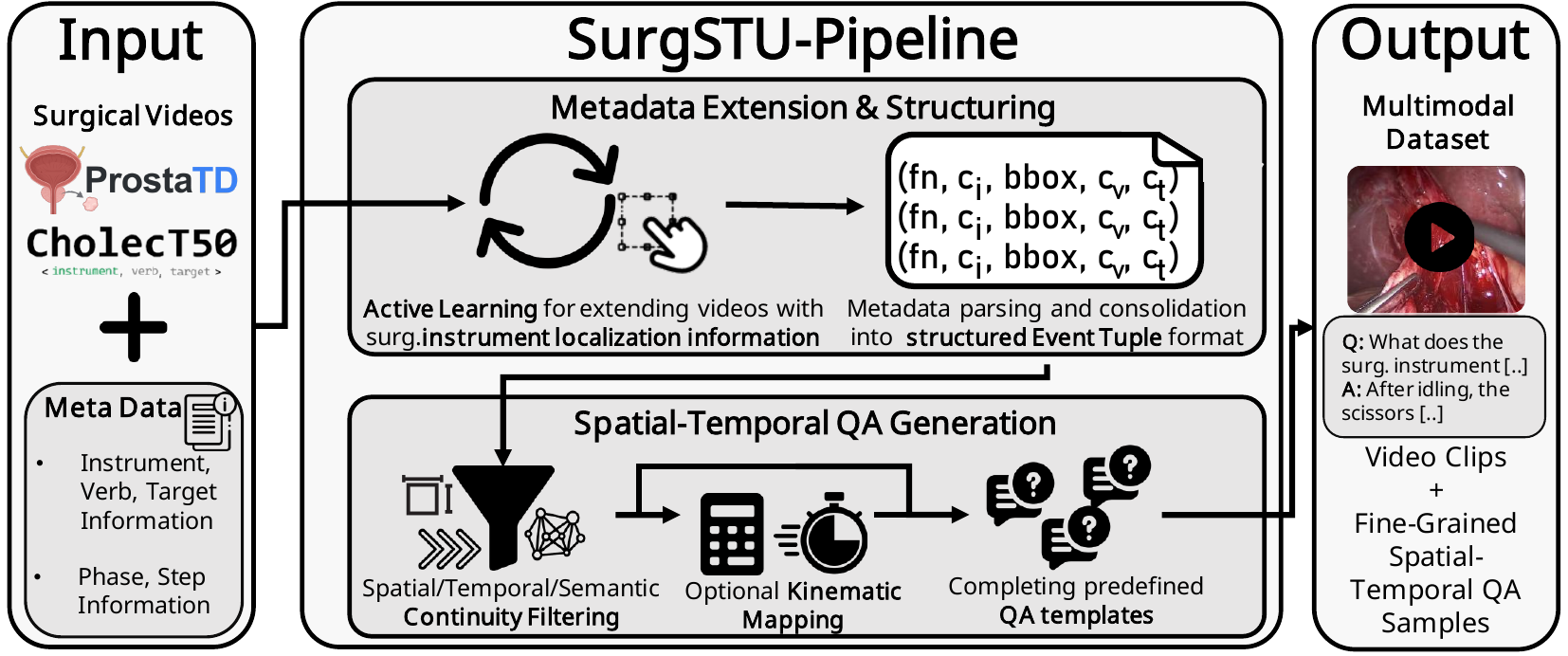}
    \caption{Overview of the proposed SurgSTU-Pipeline. The pipeline is composed of two primary steps: Metadata Extension $\&$ Structuring and Spatial-Temporal QA Generation. First, an active learning paradigm extends raw surgical videos with instrument localizations information, which are then consolidated with existing metadata into a structured Event Tuple format. Subsequently, the Spatial-Temporal QA generation module applies spatial and temporal continuity filtering to these tuples to mitigate annotation noise. Finally, the filtered sequences are processed through predefined task templates to deterministically generate a multimodal dataset comprising diverse, fine-grained spatial-temporal question-answer pairs.}
    \label{fig:dataset_curation_and_qa_generation}
\end{figure*}

\subsection{Multimodal Datasets in Surgery}

The development of surgical vision-language models is closely connected to the generation of multimodal datasets.
Early datasets such as EndoVis-18-VQLA provided question-answer pairs, including spatial grounding tasks, by incorporating bounding box annotations and instrument-tissue interactions. \cite{bai2023surgical}.
However, these datasets were confined to static frames, limited in scale, and lacked the diversity of complex natural language.
The SVL-Pretrain dataset assembled 1.3K videos and over 25K clip-text pairs by transcribing lecture audio, providing rich language supervision without manual labeling \cite{yuan2025learning}.
HecVL expanded this paradigm by organizing the SVL dataset into hierarchical levels to better represent the structure of surgical activities \cite{yuan2024hecvl}.
Nevertheless, these datasets sometimes suffered from transcription noise or a temporal misalignment between audio and visual actions.
SurgLaVi established a massive surgical dataset of nearly 240K clip-caption pairs across more than 200 procedure types \cite{PEREZ2026103982}.
It utilizes a fully automated pipeline to generate semantically coherent annotations at coarse, mid, and fine temporal granularities.
Similarly, the SVU-31K dataset was designed for multi-grained video comprehension, containing over 31K video-instruction pairs augmented by GPT-4 \cite{wang2025surgvidlm}.
For static image understanding, Surg-396K recently provided 396K multimodal instruction-following annotations derived from existing datasets, supporting conversational templates like Grounding QA and Region Based QA \cite{wang2025endochat}. \\
However, existing surgical multimodal datasets lack interconnected fine-grained spatial-temporal tasks and information.
Current benchmarks treat spatial grounding and temporal reasoning as disjoint challenges or lack the explicit fine-grained spatial tracking required to fully ground complex surgical interactions over time.

\section{SurgSTU-Pipeline}
With the SurgSTU-Pipeline, we aim to reliably generate surgical datasets for fine-grained spatial-temporal multimodal understanding.
The SurgSTU-Pipeline is visualized in Figure \ref{fig:dataset_curation_and_qa_generation} and consists of the Metadata Extension $\&$ Structuring step and the Spatial-Temporal QA Generation step.
The following subsections describe the two steps of the SurgSTU pipeline and the evaluation metrics used to accurately assess the capabilities of vision-language models.

\subsection{Metadata Extension $\&$ Structuring}

The Metadata Extension $\&$ Structuring step involves extending existing surgical video datasets with dense spatial annotations in the form of bounding-box localizations for all visible surgical instruments.
This is achieved through an iterative active learning paradigm, where over multiple rounds, a subset of frames are manually annotated with precise bounding boxes.
Subsequently, an object detection model infers instrument localizations across the remaining unannotated data. To maximize annotation efficiency and model accuracy, frames where the model exhibits low prediction probability scores are specifically prioritized for manual review and correction in the subsequent round. \\
Once the dense localizations are finalized, the raw videos and other available annotations, e.g. instrument-tissue interactions, are parsed and consolidated into a clean, structured "Event Tuple" format.
This structuring ensures that the downstream Spatial-Temporal QA Generation process can efficiently and accurately query spatial and temporal relationships within the surgical scene. \\
For example, in the case of available instrument-tissue interaction information, the "Event Tuple" can be defined as $\mathcal{T} = \langle \mathbf{fn}, \mathbf{c_i}, \mathbf{bbox}, \mathbf{c_v}, \mathbf{c_t} \rangle$, where $\mathbf{fn}$ represents the timestamp of the corresponding video frame, $\mathbf{bbox} = [x_1, y_1, x_2, y_2]$ denotes the normalized bounding box coordinates of the surgical instrument, $\mathbf{c_v}$ is the verb class describing the instrument's action and $\mathbf{c_t}$ is the target class representing the object of interaction.
Instances where specific interaction data is unavailable are encoded as null.

\subsection{Spatial-Temporal QA Generation}

The Spatial-Temporal QA Generation stage systematically translates structured ”Event Tuples” into diverse, natural language understanding tasks using predefined question and answer templates.
This way the fine-grained understanding of Vision-Language Models can be probed. 
An overview of predefined spatial-temporal question types as well as corresponding exemplary QA templates can be observed in more detail in Table \ref{tab:qa_samples_mixed}.\\
To ensure correctness in generating spatial-temporal task, such as spatial-temporal instrument grounding or velocity estimation, the SurgSTU-Pipeline incorporates a spatial and temporal continuity filtering.
Before any longitudinal query is formulated, the filter algorithm verifies that an instrument is present in every expected frame within the query window $[t_{start}, t_{end}]$.
It simultaneously enforces spatial continuity by rejecting segments where bounding box centroid displacements exceed a specific threshold value.
Once verified, deterministic kinematic mapping calculates Euclidean distances over specific durations to derive ground-truth speeds, which are then rigidly mapped to semantic descriptors (e.g., "stationary" or "moving actively"). \\
In the case of available surgical instrument-tissue interaction annotations, the SurgSTU-pipeline groups frame-level event tuples into continuous semantic blocks.
This ensures that the specific surgical verb and anatomical target remain identical across the entire segment. 
This enables the SurgSTU-pipeline to precisely formulate complex transition queries, such as predicting sequential actions. \\
For the generation of Multi-Choice Tasks, the SurgSTU-Pipeline procedurally generates negative options to prevent models from exploiting multi-choice tasks via simple elimination. \\
For Chain-of-Thought tasks, the SurgSTU-Pipeline applies a multi-stage logic scheme in which vision-language models must systematically articulate visual intermediate results before reaching a final conclusion.
Specifically, the generator structures the argumentation path with spatial localization first, followed by kinematic analysis and semantic interactions.

\begin{table}[htbp]
\caption{Overview of predefined spatial-temporal question types grouped by their respective core tasks. Exemplary Question (Q) and Answer (A) templates are provided for selected tasks.}
\centering
\footnotesize
\renewcommand{\arraystretch}{1.4}
\begin{tabular}{@{} p{0.32\columnwidth} p{0.64\columnwidth} @{}}
\toprule
\textbf{\small Task \& \textit{Description}} & \textbf{\small Exemplary QA Template} \\ \midrule

\multicolumn{2}{@{}l}{\textbf{\small Spatial-Temporal Instrument Grounding}} \\

\multicolumn{2}{@{}p{0.96\columnwidth}@{}}{\textbf{Trajectory Extremes:} \textit{Extract extreme spatial coordinates (e.g., ``most left position'') within a given time frame.}} \\

\textbf{Temporal Window} \newline \textit{Define start/end appearance times and associated spatial coordinates for visible instruments.} & 
\textbf{Q:} For every surgical instrument visible at normalized timestamp [fn], provide a temporal window (normalized duration of the video clip) along with the corresponding bounding box coordinates [x1, y1, x2, y2] for its initial appearance and final disappearance. \newline
\textbf{A:} [name]: Window [[start\_fn] - [end\_fn]], Start BBox $<$x1, y1, x2, y2$>$, End BBox $<$x1, y1, x2, y2$>$ (all bounding boxes normalized [0,1000]). \\

\multicolumn{2}{@{}p{0.96\columnwidth}@{}}{\textbf{Locate Instrument:} \textit{Provide spatial coordinates for a given timestamp.}} \\

\multicolumn{2}{@{}p{0.96\columnwidth}@{}}{\textbf{Frame Segmentation:} \textit{Identify the frame segment an instrument occupies (horizontal/vertical).}} \\

\textbf{Closest Instrument} \newline \textit{Identify the tool nearest to a specific point at a given timestamp.} & 
\textbf{Q:} Which surgical instrument is located closest to the coordinates $<$x1, y1, x2, y2$>$ at [fn] of the video duration? Coordinates are in normalized [0,1000] format. \newline
\textbf{A:} The [closest\_name] is closest. \\ \midrule

\multicolumn{2}{@{}l}{\textbf{\small Referred Instrument-Interaction Captioning}} \\

\multicolumn{2}{@{}p{0.96\columnwidth}@{}}{\textbf{Sequential Actions:} \textit{Predict the immediate next action given the current instrument's action.}} \\

\multicolumn{2}{@{}p{0.96\columnwidth}@{}}{\textbf{Action/Movement Status:} \textit{Determine the ongoing verb/action for a given time frame.}} \\

\textbf{Target Interaction} \newline \textit{Identify the specific anatomical target a given instrument interacts with.} & 
\textbf{Q:} With what target is the surgical instrument [name] interacting from [t1] of the video duration to [t2]? \newline
\textbf{A:} The [name] is interacting with the \textbf{[target]}. \\

\multicolumn{2}{@{}p{0.96\columnwidth}@{}}{\textbf{Instrument Identification:} \textit{Name a tool based on its bounding box.}} \\ \midrule

\multicolumn{2}{@{}l}{\textbf{ \small Spatial-Temporal Relation Comparison}} \\

\textbf{Relative Position} \newline \textit{Describe static spatial relationships between tools.} & 
\textbf{Q:} Describe the relative position of the surgical instrument [name1] with respect to the surgical instrument [name2] at [fn] of the video duration. \newline
\textbf{A:} The [name1] is located to the [right/left] and [below/above] the [name2]. \\

\multicolumn{2}{@{}p{0.96\columnwidth}@{}}{\textbf{Relative Change:} \textit{Describe dynamic spatial distance changes over time.}} \\

\multicolumn{2}{@{}p{0.96\columnwidth}@{}}{\textbf{Interaction Comparison:} \textit{Compare the interaction activities of two instruments.}} \\ \bottomrule

\end{tabular}
\label{tab:qa_samples_mixed}
\end{table}

\subsection{Evaluation Metrics}
In order to assess vision language models on fine-grained spatial-temporal surgical video understanding, it is necessary to extend standard n-gram matching metrics (e.g., BLEU, ROUGE) or generic text-to-text semantic similarity scores.
Such metrics often fail to capture the precision required for spatial localizations, temporal boundaries, and specific surgical state interactions.
To address this, we introduce a customized evaluation protocol for the SurgSTU dataset that directly parses the model's generated text to compute deterministic, spatial and temporal metrics across the core tasks. \\
\textbf{MultiChoice Metrics:} 
For tasks requiring categorical reasoning or discrete surgical decision-making, we evaluate using the accuracy score by parsing the model's generated answer for the chosen option letter to compare it with the ground truth answer. \\
\textbf{Spatial-Temporal Metrics:} 
Another contribution lies in the evaluation of spatial-temporal tasks.
Instead of scoring textual overlap, our evaluation pipeline extracts predicted bounding boxes and timestamps from the generated text.
For spatial grounding tasks, we calculate the standard Intersection over Union (IoU) and spatial error using normalized center point distances between the predicted and ground-truth bounding boxes.
For temporal-based queries, e.g., determining the active window of a surgical instrument, we calculate the temporal error. 
For coherent spatial-temporal tasks, the model must simultaneously predict spatial coordinates and their corresponding timestamps.
We define the composite spatiotemporal error $E_{st}$ as the average Euclidean distance between the \textit{normalized} temporal errors $\Delta \tilde{t}_i$ and \textit{normalized} spatial errors $\Delta \tilde{s}_i$ of the two endpoints ($i \in \{1, 2\}$).
\begin{equation}
    _{ST} = \frac{1}{2} \sum_{i=1}^{2} \sqrt{ (\Delta \tilde{t}_i)^2 + (\Delta \tilde{s}_i)^2 }
\end{equation}
For spatial-temporal subtasks involving velocity or relative motion, we extract specific numeric predictions (min, max, mean speeds) and calculate the relative absolute error.
This numeric score is weighted alongside a categorical description accuracy, e.g., classifying movement as "stationary" vs. "active") \\
Evaluating referred instrument-interaction captioning requires precise semantic understanding of the surgical scene.
Our pipeline employs synonym-aware parsing to map diverse generated terminology (e.g., standardizing "clipping" to "clip") to canonical labels.
For straightforward queries, we compute exact-match accuracy on the extracted entities, such as surgical actions, target tissues, and instruments.
Furthermore, tasks requiring logical reasoning, such as comparing the states of two distinct instruments, are evaluated via weighted composite scores.
These composite metrics demand both the correct logical conclusion (e.g., a "yes/no" answer) and the accurate identification of the underlying entities.

\begin{table*}[t]
\centering
\caption{Results on the SurgSTU test set grouped by the various core tasks for fine-grained spatial-temporal multimodal understanding in surgical videos. The scores represent the micro-average for each core task, calculated using the corresponding subtasks. ST Grounding: Spatial-Temporal Instrument Grounding, Ref.Int. Captioning: Referred Instrument-Interaction Captioning, ST Rel. Comp: Spatial-Temporal Relation Comparison,  MC: Multi-Choice.}
\label{tab:core_task_results}
\resizebox{\textwidth}{!}{
\begin{tabular}{l c c c c c c c}
\toprule
\textbf{Model} & \textbf{ST Grounding} & \textbf{Ref.Int. Captioning} & \textbf{Velocity Est.} & \textbf{ST Rel. Comp.} & \textbf{MC Counting} & \textbf{MC Existence} & \textbf{MC Class} \\
& (Primary $\uparrow$) & (Primary $\uparrow$) & (Primary $\uparrow$) & (Primary $\uparrow$) & (Primary $\uparrow$) & (Primary $\uparrow$) & (Primary $\uparrow$) \\
\midrule
\multicolumn{8}{c}{\textit{Without In-Context Learning}} \\
\midrule
\textbf{Generalist VLMs} & & & & & & & \\
\quad Qwen3-VL 2B & 13.72 & 21.56 & 12.28 & 23.68 & 44.29 & 55.39 & 35.87 \\
\quad Gemini-3.1 Flash-Lite & 24.98 & 27.05 & 43.24 & 18.22 & 50.46 & 59.43 & 62.49 \\
\addlinespace
\textbf{Surgical-Domain VLM} & & & & & & & \\
\quad EndoChat & 1.19 & 17.90 & 0.00 & 12.96 & 0.00 & 0.00 & 0.00 \\
\midrule
\multicolumn{8}{c}{\textit{With In-Context Learning}} \\
\midrule
\textbf{Generalist VLMs} & & & & & & & \\
\quad Qwen3-VL 2B & 36.72 & 25.01 & 53.65 & 42.08 & 53.86 & 61.99 & 26.32 \\
\quad Gemini-3.1 Flash-Lite & 45.01 & 40.01 & 47.16 & 37.68 & 60.54 & 64.97 & 63.28 \\
\addlinespace
\textbf{Surgical-Domain VLM} & & & & & & & \\
\quad EndoChat & 5.56 & 21.55 & 0.00 & 23.95 & 27.91 & 6.31 & 21.60 \\
\midrule
\multicolumn{8}{c}{\textit{SurgSTU Fine-Tuned}} \\
\midrule
\quad Qwen3-VL 2B & \textbf{60.88} & \textbf{61.93} & \textbf{70.42} & \textbf{51.90} & \textbf{98.42} & \textbf{85.76} & \textbf{90.20} \\
\bottomrule
\end{tabular}
}
\end{table*}


\section{Experiments}

In this section, we present an evaluation to benchmark the current capabilities of Vision Language Models for fine-grained spatial-temporal multimodal understanding tasks in surgery.
To do so, we apply the SurgSTU-Pipeline on the two largest publicly available surgical datasets for surgical triplet recognition, i.e., CholecT50 and ProstaTD \cite{nwoye2022rendezvous, chen2026prostatd}, to generate the SurgSTU dataset. 
Using the generated dataset, we aim to quantify the zero-shot performance of both state-of-the-art generalist foundation models and a surgical-specific VLM, i.e., Endochat, on fine-grained spatial-temporal understanding tasks.
Second, we investigate the impact of test-time interventions, specifically In-Context Learning (ICL) \cite{dong2024survey}, to determine if explicitly providing domain-specific examples can overcome the surgical domain gap for generalist models.
Lastly, we establish a benchmark by fully fine-tuning a generalist VLM (Qwen3-VL 2B) on the SurgSTU training set.

\subsection{Generated SurgSTU Dataset}

Both utilized datasets, i.e., CholecT50 and ProstaTD, contain information on fine-grained surgical instrument-tissue interactions.
These interactions are defined using triplets consisting of instrument, verb, and tissue/target.
Since the existing annotations are strictly image-based and sampled at a sparse one fps rate, our first step was to reconstruct continuous video sequences.
We achieved this by leveraging the original 30 fps videos from their respective origin datasets, Cholec80 and GraSP \cite{endonet, ayobi2024pixelwise, ayobi2023matis}.
To densely propagate the instrument-tissue interaction annotations across the temporal dimension, we interpolated the sparse 1 fps annotations by broadcasting the label of each annotated frame to all neighboring frames within a $\pm$0.5-second temporal window.
Subsequently, we divide the long videos we have collected into video clips with a length of between 20 and 30 seconds.
Finally, we feed the video clips and associated metadata into the SurgSTU pipeline, which generates the SurgSTU training and test dataset.
This process yields 5,058 video clips with 115,118 QA samples for training, and 1,653 video clips with 33,330 QA samples for testing. No videos from the same procedure appear in both the training and test sets.

\subsection{Experimental Setup}

As generalist foundation models, we evaluate the open-weight Qwen3-VL 2B \cite{bai2025qwen3} and Google's closed-source Gemini-3.1 Flash-Lite model \cite{geminiteam2023gemini}.
Furthermore, we include a publicly available multimodal model from the surgical domain, i.e., EndoChat \cite{wang2025endochat}. \\
For domain-specific fine-tuning of Qwen3-VL 2B, we employ a two-phase curriculum strategy.
Phase 1 aligns the multi-modal projector using spatial image-text pairs with a learning rate of 2e-4 for one epoch.
Phase 2 unfreezes the projector and applies Low-Rank Adaptation (LoRA \cite{hu2022lora}, $r=64$, $\alpha=128$) to the LLM backbone using spatial-temporal video data from the SurgSTU training set with a learning rate of 2e-5 for two epochs.
The Qwen3-VL 2B is trained on a single NVIDIA RTX 4090. \\
To evaluate the impact of surgical domain specific vocabulary on generalist models, we test two different prompting paradigms
In the Zero-Shot setup, models are simply provided the video and a corresponding question.
In the In-Context Learning (ICL) setup, we extend the prompt with domain-specific information, i.e., a list of candidate surgical instruments, verbs, and targets. 
Additionally, we dynamically retrieve one successful exemplar from the SurgSTU training split that matches the category of the test question. 
This is done to guide the model in terms of the expected response format and the spatial-temporal reasoning style.

\subsection{Results}
Table \ref{tab:core_task_results} presents the quantitative evaluation of the examined Vision-Language Models on the SurgSTU test set. \\
In the baseline setting without In-Context Learning (ICL), generalist foundation models struggle with complex spatial-temporal reasoning, underscoring the difficulty of the benchmark. 
Among these, Gemini-3.1 Flash-Lite achieves the highest zero-shot performance, scoring 24.98\% in Spatial-Temporal Grounding and 43.24\% in Velocity Estimation.
Conversely, the surgical-specific model EndoChat exhibits severe limitations in this setting, failing entirely in Velocity Estimation and all Multi-Choice (MC) tasks. \\
The application of test-time ICL interventions substantially enhances the results of the examined models, demonstrating that providing domain-specific exemplars can mitigate the surgical domain gap.
With ICL, Qwen3-VL 2B demonstrates a performance leap, improving its Spatial-Temporal Grounding score from 13.72\% to 36.72\%, and its Velocity Estimation score from 12.28\% to 53.65\%.
Gemini-3.1 Flash-Lite also shows consistent gains across most tasks, notably reaching 45.01\% in Spatial-Temporal Grounding tasks and 40.01\% in Referred Instrument-Interaction Captioning tasks.
EndoChat also benefits from ICL, showing modest gains in spatial-temporal relation comparison (12.96\% to 23.95\%) and partially recovering from complete failure on Multi-Choice tasks. \\
Finally, the SurgSTU Fine-Tuned Qwen3-VL 2B shows highest performance across all core tasks.
By fully aligning the model to the surgical domain using the SurgSTU training dataset, it outperforms both zero-shot and ICL baselines, achieving 60.88\% in Spatial-Temporal Grounding tasks and 70.42\% in Velocity Estimation.
Furthermore, the fine-tuned model reaches highest accuracies of 98.42\% in Multi-Choice Counting and 90.20\% in Multi-Choice Class.

\begin{figure*}[t]
    \centering
    \includegraphics[width=\textwidth]{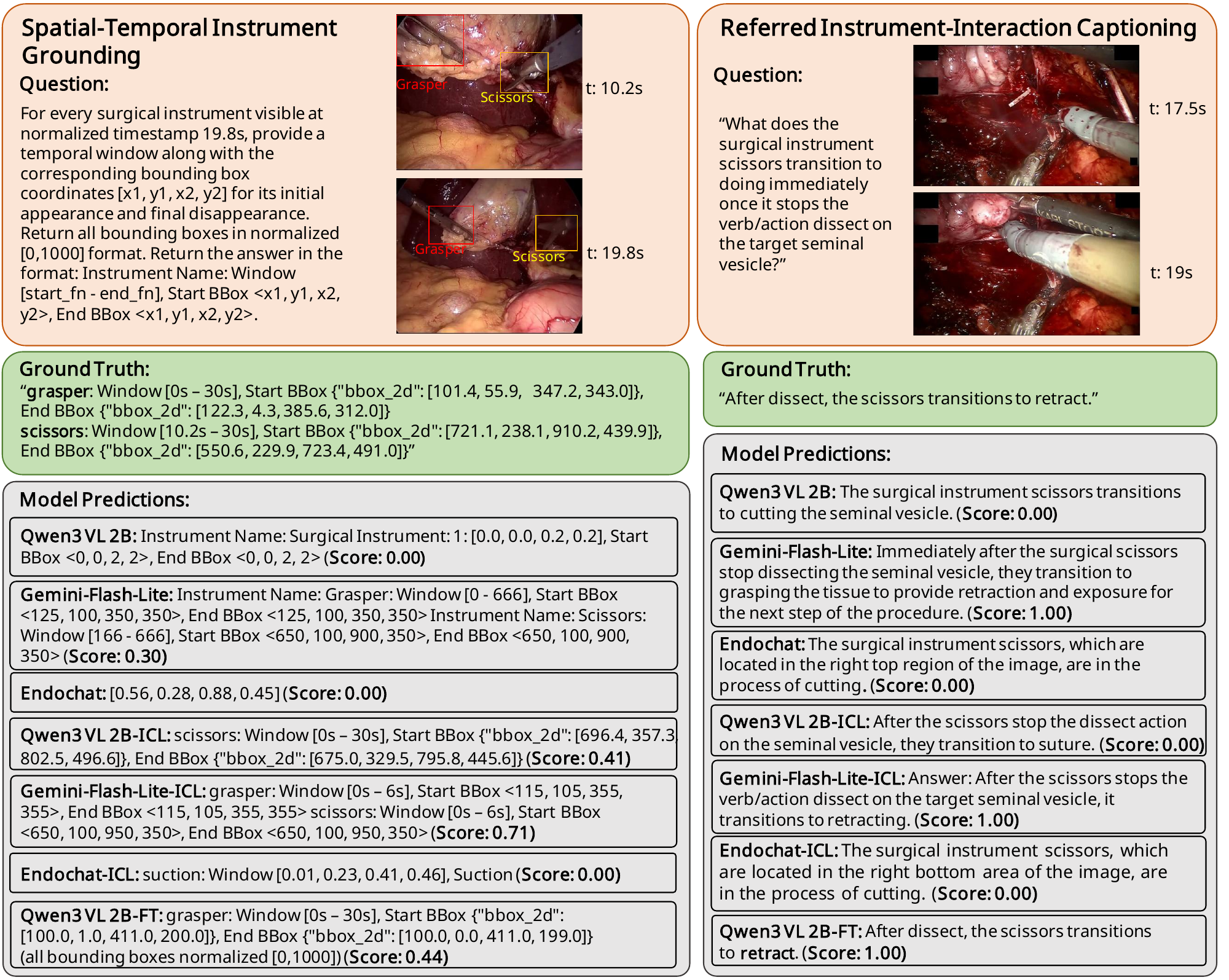}
    \caption{Qualitative results for Spatial-Temporal Instrument Grounding and Referred Instrument-Interaction Captioning tasks on the generated SurgSTU dataset. Model predictions from Qwen3 VL 2B, Gemini-Flash-Lite, and EndoChat are compared across Zero-Shot, In-Context Learning (ICL), and Fine-Tuned (FT) configurations. Result scores are attached at the back of each respective Model Prediction.}
    \label{fig:results}
\end{figure*}

\section{Discussion and Conclusion}

In this work, we introduce the SurgSTU-Pipeline, a deterministic approach to enriching existing surgical video datasets for fine-grained spatial-temporal understanding.
By applying spatial and temporal continuity filtering, we generate the SurgSTU dataset from publicly available datasets CholecT50 and ProstaTD, comprising 6,711 video clips densely annotated with 150k robust multimodal question-answer samples.
Our pipeline effectively addresses the limitations of coarse, static image-level benchmarks by explicitly capturing complex, interleaved spatial-temporal dynamics. \\
Our evaluation demonstrates that while state-of-the-art generalist Vision-Language Models struggle in zero-shot settings, domain-specific fine-tuning (e.g., with Qwen3-VL 2B) yields the highest performance across all spatial-temporal tasks.
However, the fine-tuned Qwen3-VL model might overfit to the specific visual distributions of cholecystectomy and prostatectomy procedures in the SurgSTU dataset.
An important next step is evaluating how these spatial-temporal capabilities generalize to unseen surgical disciplines with novel instruments.
Furthermore, test-time interventions via In-Context Learning (ICL) leads to notable quantitative improvements for the generalist models.
This boost is also attributable to the fact that domain-specific exemplars enable the models to generate responses more robustly in the output formats required for calculating our deterministic evaluation metrics. 
While standard text-overlap metrics might fail to capture fine-grained information, e.g., an exact instrument's location, our deterministic evaluation harshly penalizes models that fail to output the expected coordinate formats. \\
The surgical domain-specific model, EndoChat, exhibits significant limitations.
Because EndoChat relies on aggregating static image-level predictions, continuous temporal reasoning tasks, such as Velocity Estimation, are inherently difficult. Additionally, the model struggles to process the structured Multi-Choice formats.
We hope to conduct experiments with video-based generative VLMs from the surgical domain as soon as they are made publicly available. \\
From a dataset curation perspective, the SurgSTU-Pipeline demonstrates good scalability for specific task categories and functions as an extension to fully scalable datasets like SVL and SurgLaVi.
By leveraging bounding box metadata, it is possible to automatically generate a vast quantity of Spatial-Temporal Instrument Grounding tasks.
However, generating tasks that require complex semantic knowledge, such as Referred Instrument-Interaction Captioning, remains bottlenecked by the need for dense and highly costly manual annotations. \\
Additionally, despite the implementation of temporal and spatial continuity filtering within our pipeline, a fraction of the generated QA samples may still contain inaccuracies.
This noise propagates from errors in the initial ground-truth data from which interactions are sampled, imperfect bounding box localizations introduced during the active learning phase, or misalignment caused by the temporal interpolation of sparsely annotated frames. \\
Despite these challenges, the SurgSTU-Pipeline enables the deterministic generation of surgical datasets for fine-grained spatial-temporal multimodal understanding, addressing a critical gap in existing surgical multimodal benchmarks.

{
    \small
    \bibliographystyle{ieeenat_fullname}
    \bibliography{main}
}


\end{document}